\documentclass[runningheads]{llncs}
\usepackage[T1]{fontenc}
\usepackage{amsmath,amssymb,amsfonts}
\usepackage{algorithmic}
\usepackage{graphicx}
\usepackage{textcomp}
\usepackage{xcolor}
\usepackage{soul,bm}

\title{The Joint Weighted Average (JWA) Operator}

\author{Stephen B. Broomell\inst{1} \and Christian Wagner\inst{2}}
\institute{GRID -- Global Risk and Individual Decisions Laboratory, \\Purdue University, West Lafayette IN 47907, USA,\\
        \email{Broomell@purdue.edu}
        \and
        LUCID -- Lab for Uncertainty in Data and Decision Making, \\University of Nottingham, Nottingham, UK, \\
        \email{Christian.Wagner@nottingham.ac.uk}}

\date{April 2024}

\begin{document}

\maketitle

\begin{abstract}
Information aggregation is a vital tool for human and machine decision making in the presence of uncertainty. Traditionally, approaches to aggregation broadly diverge into two categories, those which attribute a worth or weight to information sources and those which attribute said worth to the evidence arising from said sources. The latter is pervasive in the physical sciences, underpinning linear order statistics and enabling non-linear aggregation. The former is popular in the social sciences, providing interpretable insight on the sources. While prior work has identified the need to apply both approaches simultaneously, it has yet to conceptually integrate both approaches and provide a semantic interpretation of the arising aggregation approach. Here, we conceptually integrate both approaches in a novel joint weighted averaging operator. We leverage compositional geometry to underpin this integration, showing how it provides a systematic basis for the combination of weighted aggregation operators--which has thus far not been considered in the literature. We proceed to show how the resulting operator systematically integrates a priori beliefs about the worth of both sources and evidence, reflecting the semantic integration of both weighting strategies. We conclude and highlight the potential of the operator across disciplines, from machine learning to psychology.

\end{abstract}


\section{Introduction}

Whether it be for a computerized system or for human decision making, evidence derived from information sources serves as decision inputs, and the discord between pieces of evidence can often be addressed by considering multiple of said sources. For example, museums commonly use multiple thermometers to ensure the maintenance of an ideal temperature reliably, while expert committees rely on the presence of multiple experts to address discord and individual knowledge-gaps.

Two different approaches are commonly used to aggregate information. The first approach is a linearly weighted average (LWA), which underlies many approaches to information aggregation for wisdom of crowds \cite{budescu2015identifying,davis2014crowd,davis2015composition,huang2022hypothesis}. 
The second approach is operationalized by ordered combinations of the evidence or judgment arising from the information sources. Flexible non-linear operators such as linear order-statistics have been applied with great success in applications from cyber-security to data-fusion \cite{MurrayChoquet2021,MILLER2017_CyberSec}. 
For example, the Order Weighted Average (OWA), \cite{yager81} retains much of the simplicity and interpretability of the LWA but allows for non-linear combinations of evidence. To achieve this, the OWA focuses \emph{not} on the notion of worth arising from each source, but instead on the worth associated with the actual information, i.e. the \emph{evidence} arising from each given source at a given point in time. 

Prior literature has developed several operators that attempt to combine the advantages of the OWA and the LWA. These include the Weighted OWA (WOWA) \cite{torra1996weighted,torra2009wowa} 
the Hybrid Weighted Average (HWA) \cite{xu2003overviewAgg},  
the Ordered Weighted Averaging – Weighted Averaging (OWAWA) \cite{merigo2009wOWA}, 
and the Standard Deviation OWA (SDOWA) \cite{SDOWA_Cardin2021}. 
The former two operators have limitations to their behavior, as outlined by Llamazares \cite{llamazares2011generalizations}, and the latter two operators are based on the partial consideration of the OWA and LWA (i.e. each is used to a different degree), rather than the \emph{joint} consideration of them. We show that all prior operators lack a mathematical framework that provides a semantic justification and definition for how the LWA and OWA should be combined.

In this paper, we introduce such a mathematical framework for creating joint weights and explore the nature of a joint focus on source-and-evidence specific worth of information. In Section 2 we define each of the extant aggregation operators. In Section 3 we discuss the assumptions and properties of information sources and evidence that would lead a decision maker to apply LWA, OWA, or some combination of the two for their information aggregation needs, showing inconsistent behavior in each of the existing operators. Next, we introduce compositional geometry \cite{Aitchison1982,smithson2022compositional} 
and discuss how this mathematical framework accounts for prior inconsistencies, defining precisely joint weights that create aggregates based on both source and evidence. In Section 4 we introduce our new aggregation operator--the Joint Weighted Average (JWA)--and in Section 5 we provide an illustrative experiment. Finally, Section 6 is the discussion and conclusion.

\section{Preliminaries}

\subsection{Basic Aggregation Operators and their Properties}

Prior work has established properties of aggregation functions that define their behavior (see \cite{SDOWA_Cardin2021,llamazares2011generalizations}). Some of these properties are defined because they are desirable, but some are defined in order to establish logical bounds on the behavior of an aggregator. Let $F$ be a function that takes in a vector $\bm{x} \in R^n$ of observations and generates an aggregate. The function $F$ may or may not have the following properties:
\begin{itemize}
    \item[1.] \textbf{Compensativeness.} For all $\bm{x} \in R^n$, $min(\bm{x}) \le F(\bm{x}) \le max(\bm{x})$.
    \item[2.] \textbf{Idempotency.} For all $x \in R$, $F(x,\ldots,x) = x$.
    \item[3.] \textbf{Homogeneity of degree 1.} For all $\bm{x} \in R^n$ and $\lambda > 0$, $F(\lambda\bm{x}) = \lambda F(\bm{x})$.
    \item[4.] \textbf{Monotonicity.} For all $\bm{x}, \bm{y} \in R^n$, if $x_i \ge y_i$ for all $i \in \{1,\ldots,n\}$ then $F(\bm{x}) \ge F(\bm{y})$.
    \item[5.] \textbf{Commutativity.} For all $\bm{x} \in R^n$ and for all permutations of $\bm{x}$ given by $\bm{x}_\pi$, $F(\bm{x}_\pi) = F(\bm{x})$.
\end{itemize}

\subsubsection{The Linear Weighted Average (LWA)}
Let $\bm{w}$ be a vector of weights such that $\bm{w} \in [0, 1]^n$ and $\sum_{i=1}^nw_i = 1$. The LWA is a linear combination of the evidence provided by each information source ($x_i$) given by,
\begin{equation}
LWA(\bm{x}) = \sum^n_{i=1}w_ix_i.
\end{equation}
The LWA satisfies properties 1 - 4, and conditionally satisfies property 5 if and only if $w_i = 1/n$ for all $i \in \{1,\ldots,n\}$.

\subsubsection{The Order Weighted Average (OWA) \cite{yager81}}
Let $\bm{v}$ be a vector of weights such that $\bm{v} \in [0, 1]^n$ and $\sum_{i=1}^nv_i = 1$. Let $\pi()$ be a permutation function such that $x_{\pi(1)}
\geq ... \geq x_{\pi(n)}$. The OWA is an ordered combination of the evidence provided by each information source ($x_i$) given by,
\begin{equation}
OWA(\bm{x}) = \sum^n_{i=1}v_ix_{\pi(i)}    
\end{equation}
The OWA satisfies properties 1 - 5.

\subsection{Combination Aggregation Operators and their Properties}

Aggregation operators that combine the LWA and OWA can introduce additional complexity. Their behavior has been understood in prior work by an additional set of properties that define their relationship to the LWA and OWA as outlined by \cite{SDOWA_Cardin2021,llamazares2011generalizations,torra1996weighted}. 
\begin{itemize}
    \item[6.] \textbf{Collapsing.} For all $\bm{x} \in R^n$, if $w_1 = \ldots = w_n$ then, $F(\bm{x}) = OWA(\bm{x})$ and if $v_1 = \ldots = v_n$ then $F(\bm{x}) = LWA(\bm{x})$.
    \item[7.] \textbf{Internal Boundedness.} For all $\bm{x} \in R^n$, $min(OWA(\bm{x}),LWA(\bm{x})) \le F(\bm{x}) \le max(OWA(\bm{x}),LWA(\bm{x}))$.
    \item[8.] \textbf{Transitivity of Equality.} For all $\bm{x} \in R^n$, if $OWA(\bm{x}) = LWA(\bm{x}) = k$ then $F(\bm{x})= k$.
\end{itemize}

\subsubsection{The Weighted OWA (WOWA) \cite{torra1996weighted}}
Let $\bm{w}$ be the vector of weights defined by LWA and $\bm{v}$ be the vector of weights defined by OWA. Let $\pi()$ be the permutation function defined by OWA. Let the function s() interpolate between points $\left(i/n, \sum_{j = 1}^iv_j\right)$ together with $(0, 0)$. A new weight vector $\bm{\omega}$ is computed as,
\begin{equation}
\omega_i = s\left(\sum_{j = 1}^iw_{\pi(j)}\right) - s\left(\sum_{j = 1}^{i-1}w_{\pi(j)}\right).
\end{equation}
The WOWA 
is an ordered combination of the evidence provided by each information source ($x_i$) given by,
\begin{equation}
WOWA(\bm{x}) = \sum^n_{i=1}\omega_ix_{\pi(i)}.
\end{equation}
The WOWA satisfies properties 1 - 4 and 6 \cite{llamazares2011generalizations}. It conditionally satisfies property 5 if and only if $w_i = 1/n$ for all $i \in \{1,\ldots,n\}$ and it collapses to the OWA.

\subsubsection{The Hybrid Weighted Average (HWA) \cite{xu2003overviewAgg}}
Let $\bm{w}$ be the vector of weights defined by LWA and $\bm{v}$ be the vector of weights defined by OWA. Let $\pi()$ be a permutation function such that $nw_{\pi(1)}x_{\pi(1)} \geq \ldots \geq nw_{\pi(n)}x_{\pi(n)}$.  The HWA 
is an ordered combination of the evidence provided by each information source ($x_i$) given by,
\begin{equation}
HWA(\bm{x}) = \sum^n_{i=1}v_i(nw_{\pi(i)}x_{\pi(i)}).
\end{equation}
The HWA satisfies properties 3, 4, and 6  \cite{llamazares2011generalizations}. It conditionally satisfies properties 1 and 2 if and only if it collapses to the LWA or OWA, and property 5 if and only if it collapses to OWA.

\subsubsection{Ordered Weighted Average - Weighted Average (OWAWA) \cite{merigo2009wOWA}}
A simple method for combining the LWA and OWA is to generate a weighted combination of the two aggregators. The OWAWA operator is given by,
\begin{equation}
OWAWA(\bm{x}) = \alpha LWA(\bm{x}) + (1 - \alpha)OWA(\bm{x}).
\end{equation}
The parameter $\alpha \in [0, 1]$ defines the relative contribution of the LWA and OWA to the final aggregate. The OWAWA satisfies properties 1 - 4, 7, and 8 \cite{SDOWA_Cardin2021}.
It conditionally satisfies property 5 if and only if $w_i = 1/n$ for all $i \in \{1,\ldots,n\}$ or $\alpha = 0$.

\subsubsection{Standard Deviation OWA (SDOWA) \cite{SDOWA_Cardin2021}}
Let the function $sd()$ return the standard deviation of a vector. Let the function $G()$ be defined as, 
\begin{equation}
G(\bm{w},\bm{v}) = \frac{sd(\bm{w})}{sd(\bm{w}) + sd(\bm{v})}.  
\end{equation}
The SDOWA operator 
is given by,
\begin{equation}
\begin{split}
SDOWA(\bm{x}) = G(\bm{w},\bm{v})LWA(\bm{x})+ \\(1-G(\bm{w},\bm{v}))OWA(\bm{x}). 
\end{split}
\end{equation}
The SDOWA satisfies properties 1 - 4 and 6 - 8 \cite{SDOWA_Cardin2021}. It conditionally satisfies property 5 if and only if $w_i = 1/n$ for all $i \in \{1,\ldots,n\}$.

\section{Theoretical Context}

In order to create an operator that fully integrates the LWA and the OWA, it is valuable to take a step back and conceptually understand the properties of the evidence we wish to aggregate and the desired outcome of integrating two aggregators. We start with a conceptual example where linear and order weighting both apply, and show how each aggregator defined above fails to produce a coherent aggregate.

Consider the problem of estimating how much product to produce, using judgments from 3 experts. First, we may have some prior knowledge of the relative performance of each expert such that we assign weights \{0.60, 0.30, 0.10\} to experts 1, 2 and 3 respectively. These weights represent an LWA, the goals of which are to give weight to the experts (sources) that matches their relative reliability or validity. Second, we may have a desire to avoid underproduction of our product such that we assign ordered weights \{0.45, 0.50, 0.05\} to the highest, middle, and lowest judgment produced by the experts. These weights represent an OWA, the goals of which are to consider the middle judgment most strongly along with the highest judgment, and put very little weight on the lowest judgment. 

It is clear that beyond the linear and non-linear characteristics of the LWA and OWA, both operators operationalize fundamentally different foci, with the LWA \emph{focusing} on the \emph{worth} of the information sources, and the OWA focusing on the \emph{worth} of the evidence or judgment contributed by the sources. Another way of looking at this is that the LWA operationalizes a strategy aligned with the sources, for example here: combine considering the expertise of the experts; while the OWA operationalizes a strategy aligned with the evidence, for example here: avoid under-production. 
If we apply the OWA (or the LWA) in isolation, we implicitly assume that all experts (or evidence orderings) are equally valid--we apply aggregation based on one \emph{or} the other strategy--not both. However, we wish to apply both sets of weights to aggregate the expert judgments to account for both strategies.

Table \ref{tab:example} displays aggregation results for each of the operators defined in the previous section using example evidence. We display the weight that was assigned to each judgement and the final aggregates of each operator. The weights show how well each operator represents the strategies defined by the LWA and OWA weights. Similar to examples provided by\cite{llamazares2011generalizations}, the WOWA exhibits strange behavior by shifting some weight from expert 2 to expert 3 (the least valued expert and order), appearing to act against the goals of both the LWA and the OWA weights. The HWA generates weights for the evidence that do not sum to one, as such, the aggregate is outside the bounds of the data, violating compensativeness. However, the HWA weights do appear to represent both the LWA and OWA strategies, reducing the weight for expert 3's judgment (least trusted expert and order) and increasing the weight of expert 1 (most trusted expert and good order location). 

We can also illustrate the difference between a weighted combination of both operators and the joint consideration of both aggregation strategies. Based on the results in Table \ref{tab:example}, any weighted aggregation of LWA and OWA (which includes the SDOWA and OWAWA) would by definition generate an estimate in the range [66, 70], i.e. between the output produced by the OWA and LWA respectively. Applying more weight to LWA will lead to a higher estimate, whereas applying more weight to OWA would lead to a lower estimate. This result is slightly counter-intuitive to the goals of the two strategies, as the OWA (focused on down-weighting the lowest judgment) leads to a lower aggregate. This result is due to the fact that the highest judgment is coming from expert 1, who is (by chance in our example) given more weight in the LWA relative to the weight assigned to the highest judgment in the OWA. Indeed, Table \ref{tab:example} shows the SDOWA is in between the LWA and OWA.

This example illustrates one primary objective for combining the LWA and OWA: to shift the relative weights applied to the evidence in such a way as to jointly represent the strategies of the linear \textit{\textbf{and}} order weights. The weights are parts of a whole, that represent relative worth. As such, each weight depends on the other weights, creating a dependency between them. The joint consideration of the two sets of weights needs mathematics that accounts for these properties of the weights. In the case of the LWA and OWA, the weights can be represented as compositions, and the mathematics and geometry of compositions can clarify the meaning behind shifting weights along with a deeper understanding of the mechanisms for computing joint weights.

\begin{table}
\addtolength{\tabcolsep}{-1pt}
\centering
\caption{Aggregation Example}
\begin{tabular}{lcccc} \hline
	&	Expert 1	\mbox{  } &	Expert 2 \mbox{  } 	&	Expert 3 \mbox{  } 	&	\\
  \hline
Evidence   &	90.00	&	50.00	&	10.00	&		\\
 \hline
\multicolumn{3}{l}{Operator weights applied to each expert} &  &	Aggregate	\\
\hspace{.5cm}LWA&	0.60	&	0.30	&	0.10	&	70.00	\\
\hspace{.5cm}OWA&	0.45	&	0.50	&	0.05	&	66.00	\\
\hspace{.5cm}WOWA&	0.45	&	0.42	&	0.14	&	62.60	\\
\hspace{.5cm}HWA&	0.81	&	0.45	&	0.02	&	95.55	\\
\hspace{.5cm}SDOWA&	0.53	&	0.40	&	0.08	&	68.02	\\
\hspace{.5cm}JWA&	0.64	&	0.35	&	0.01	&	74.94	\\
\hline
\end{tabular}
\label{tab:example}
\end{table}

\subsection{Compositional Geometry}

The aim of our paper is to enable the blending of source- and evidence-based strategies towards aggregation. Our proposed approach for this is to blend the a priori beliefs about the quality of sources and the quality of evidence. This is operationalized by blending the weights assigned to sources (via linear weights) with the weights assigned to the evidence (via order weights) into interpretable joint weights. To achieve this goal, we leverage the fact that non-negative weights that sum to one are compositional, and utilize compositional geometry to combine them.

A composition is formally defined as a collection of components that are constrained to sum to a constant for individual cases (See \cite{Aitchison1982,van2013analyzing,smithson2022compositional} for tutorials). Proportions and probabilities are examples of compositions which are constrained to sum to 1. For example, \cite{smithson2022compositional} describe how the proportion of the day spent working, relaxing, and exercising (assuming no other activities) could be defined as a composition. If more time is spent working on a given day, then there must be less time spent on relaxing and/or exercising. Therefore, components of a composition cannot be analyzed separately, because the value of one component affects the value of the others.

Compositions have their own geometry defined by a mathematical structure called a simplex.  Within the simplex, a composition is defined by each component part. For example, a component with 3 parts is defined by the set $\{a, b, c\}$ such that
$$
a \ge 0; b \ge 0; c \ge 0; \mbox{ and } a + b + c = d.
$$ 
Constraining $d$ to equal $1$ generates a standardized simplex that elegantly reflects compositions made up of proportions and probabilities. 

Because each component is dependent on the value of the remaining components, the operation of addition is slightly different in compositional geometry.
Compositional addition is known as \emph{perturbation}, and defines how the relative values of each component will change when added to another component. Given composition $\mathbf{x} = \{x_1,x_2,x_3\}$ and $\mathbf{y} = \{y_1,y_2,y_3\}$, `addition' in compositional geometry is defined using the symbol $\oplus$ as,
\begin{equation}
\mathbf{x} \oplus \mathbf{y} = \{\frac{x_1*y_1}{\sum_{k=1}^{K}{x_k*y_k}}, \ldots, \frac{x_K*y_K}{\sum_{k=1}^{K}{x_k*y_k}}\}.
\label{eq:comp_add}
\end{equation}
In compositional algebra, the additive identity is the uniform composition where $x_k = 1/K$ for all $k$, where $K$ is the number of parts, for example the number of sources in an aggregation context. 

Compositions and perturbations have the following properties,\\
(1) \textbf{scale invariance}: results do not depend on the constant sum.\\
(2) \textbf{permutation invariance}: results do not depend on the order of the parts.\\
(3) \textbf{subcompositional coherence}: results of subcompositions (any subset of component parts) do not differ from the their results when considering the full composition.

These properties make compositional geometry a reasonable mathematical representation for the linear and order based weights. The weights assigned to sources and orders are relative, meaning that the individual weights represent a part of a whole. We can use Eq. (\ref{eq:comp_add}) as an operator to blend ordered linear weights with the order weights and the resulting weights are constrained to be non-negative and sum to one.

We therefore propose \textit{joint} weights that are a \emph{perturbation} of the linear and order weights. In general, the perturbation of two quantities with different units produces a result with a joint unit. For example, \cite{van2013analyzing}
(p. 18) outline how a composition of percentages of grams of nutrients for a food item can be transformed to units of energy by adding this composition to a composition that represents the relative $kJ/g$ of each nutrient. Therefore, our joint weights are in the joint unit of source quality and evidence order.

The perturbation of two sets of weights has appeared implicitly in the aggregation literature previously, but its connection to compositional geometry was never identified. It was first introduced by \cite{engemann1996modelling} as a way of combining optimism with risk in decision making. It was also considered briefly by \cite{llamazares2011generalizations} as a method to combine order and linear weights, but was ultimately rejected on the grounds that it did not satisfy properties 7 and 8 listed above. Because the connection to compositional geometry was never made, in part because weights have not been considered as individual parts of one whole, the desirable behavior for perturbing and thus systematically and strategically integrating the linear and order weights--as described above--were unknown.

\section{The Joint Weighted Average}
\label{sec:JWA}
In this section we introduce the Joint Weighted Average (JWA) as a new aggregation operator that allows for the joint representation of independently defined linear and order weights in computed averages. We therefore propose a new type of blend of LWA and OWA with mathematical properties that differ from the prior literature attempting to combine these two weighting approaches. 

Given $n$ sources, let $\mathbf{w} = \{w_1, \ldots w_n\}$ be a set of convex source weights and let $\mathbf{v} = \{v_1, \ldots v_n\}$ be a set of convex order weights. Both $\mathbf{v}$ and $\mathbf{w}$ are compositions because their components are non-negative and sum to 1. Let $\bm{x} = \{x_1,\ldots x_n\}$ be the evidence generated by the $n$ sources. We reorder the evidence and the linear weights by the same permutation function $\pi()$, such that $x_{\pi(1)}\geq x_{\pi(2)}\geq ... \geq x_{\pi(n)}$. Finally, the ordered linear weights are perturbed by the order weights (using the compositional $\oplus$ operator), and these weights are applied to the ordered evidence to generate a weighted average given by
\begin{equation}
JWA(\bm{x}) = (\mathbf{w}_{\pi} \oplus \mathbf{v}) \bm{x}_{\pi}^T.
\end{equation}
The operations inside the parentheses represent compositional operations. We then take the inner product of the resulting composition with the data in $\mathbf{x}_{\pi}$ to calculate the weighted average. 

Returning to the example in Table \ref{tab:example}, treating the LWA and OWA's weights as compositions can resolve all the problems identified in the other operators. Given that the lowest judgment (which has the lowest priority in the OWA) is generated by expert 3 (who has the lowest priority in the LWA), a joint focus on both strategies shifts weight away from expert 3's judgment. This logic results in the JWA weights displayed in Table \ref{tab:example}, with an even larger aggregated estimate than would result from the LWA, OWA, and any weighted combination of them. The only other aggregator that displays this behavior is the HWA, but this is due to the fact that the ordering of the $\{x_1,x_2,x_3\}$ is the same as the order of $\{w_ix_1, w_2x_2,w_3x_3\}$, which is not always the case.

The mathematical properties of the JWA operator can be derived from two basic facts: (1) the vector inside the parentheses is non-negative and sums to one, making the JWA have the same properties as a weighted average and (2) that these weights are derived using well-defined compositional geometry. First, because this operator can be represented as a weighted combination of the data $\mathbf{x}_{\pi}$, it automatically satisfies properties 1 - 3. Second, within compositional geometry, the additive identity is the composition with uniform weight on each component. Therefore, if $\bm{w}_{\pi} = 1/n$ for all $i \in \{1,\ldots,n\}$ then $\bm{w}_{\pi} \oplus \mathbf{v} = \bm{v}$ and $JWA(\bm{x}) = OWA(\bm{x})$. If $\bm{v} = 1/n$ for all $i \in \{1,\ldots,n\}$ then $\bm{w}_{\pi} \oplus \bm{v} = \bm{w}_{\pi}$ and $JWA(\bm{x}) = LWA(\bm{x})$. The JWA therefore also satisfies property 6. 

\cite{llamazares2011generalizations} observed violations of monotonicity (prop. 5) from a similar perturbation of LWA and OWA. While this appears to be problematic behavior, compositional geometry reveals that this is only due to instances where the data ordering changes. In fact, JWA satisfies monotonicity if and only if $\bm{y}$ has the same ordering as $\bm{x}$, which is a more reasonable restriction. When the ordering of evidence changes, the interaction between the LWA and OWA weights should change.

Similar to the WOWA and HWA, the JWA does not satisfy properties 7 and 8. For internal boundedness (prop. 7), the results in Table 1 provide a demonstration that the JWA is not bounded by the OWA and LWA, and can in fact be less than the $min(OWA,LWA)$ or greater than the $max(OWA,LWA)$. For transitivity of equality (prop. 8), the result of $OWA(\bm{x}) = LWA(\bm{x}) = k$ can be achieved for all $\bm{x} \in R^n$ by having $\mathbf{w}_{\pi} = \mathbf{v}$, meaning the linear and order weight assigned to each element of $\mathbf{x}_{\pi}$ is the same for all $i \in \{1,\ldots,n\}$. In this case, for $JWA(\bm{x}) = k$ we require $\mathbf{w}_{\pi} = \mathbf{v} = \mathbf{w}_{\pi}\oplus\mathbf{v}$. However, this condition is only met when $\mathbf{w}_{\pi} = \mathbf{v} = \{1/n, \ldots, 1/n \}$. In all other cases, $OWA(\bm{x}) = LWA(\bm{x}) \neq JWA(\bm{x})$. 

While these behaviors violate previously desired properties, they are still mathematically sound and interpretable within compositional geometry. In fact, we find that internal boundedness (prop. 7) and transitivity of equality (prop. 8) preclude an aggregate that allows the LWA and OWA weights to interact and thus preclude the respective strategies to be blended. For example, evidence receiving weight greater (lesser) than $1/n$ from both the linear and order weights will receive even more (less) weight through perturbation, this desirable behavior and internal boundededness cannot both simultaneously hold.

Conceptualizing the JWA weights as a perturbation of two sets of relative weights requires deeper consideration of the role of weights that equal zero or one. Such values should be assigned with extreme caution. Zero weights mean that no weight will ever be assigned to that component. Therefore the LWA weights should not be assigned a zero, because a zero weight means that source's evidence should never be considered. Attempting to perturb two sets of weights where the result yields all zeros implies a logical inconsistency in the a priori weights, as the two sets of weights together suggest that none of the evidence should be used. While compositions can contain zeros, a weight of one implies all other weights are zero. Any set of weights for the LWA or OWA that contains a weight equal to one precludes the need for the JWA, as weight can no longer be perturbed in this instance.

\section{Experiments}

We provide a demonstration of the effectiveness of the JWA operator using simulation. Assume we consult $k$ sources over $n$ trials resulting in a $k$ x $n$ matrix $X$ that represents the evidence from our sources. For each trial, the aggregate of the source evidence provides a prediction $\hat{y}$ of the criterion variable $y$. 

To show the value of the JWA, we simulate a data set where the quality of the sources \textbf{and} the ordering of the data inform on the criterion variable we are trying to predict. We simulate evidence from a collection of sources, and we also simulated the criterion that we are trying to predict. This way we can directly control the statistical properties of the sources, and their evidence, for predicting $y$. We use a simple model, where the data for $X$ and $y$ are simulated from a joint normal distribution. For a normal distribution, we can set the mean, variance, and covariance of each variable. For simplicity, we set all means and variances equal to the constant 10, so that the worth of the sources is defined by the covariance structure. We set the covariance between sources ($\sigma_{x,x'}$) to the constant 2. We want the sources to have some predictive accuracy, so this is modelled by the degree to which the evidence, $X$, co-varies with the criterion, $y$. We adopt the term \emph{validity} from social sciences here to describe the correlation between our observable evidence and the criterion.

The LWA and OWA have weights that reflect the worth of the sources and the evidence in our simulation. Specifically, the worth of the sources is defined in our simulation by changing the relative validity of each source (we hold all other source attributes constant). In other words, the LWA weights directly reflect the relative validity of each source. 

To simulate a clear OWA strategy, our simulation injects changes to the interpretation of the evidence by randomly adding a positive bias (or \emph{error}) $\delta$ to the evidence generated by two of the sources. We randomly add this bias to 50\% of the trials. Since this bias is not a property of any specific source, we randomly select which two sources receive this bias independently for each trial. Therefore, our data context suggests that we should not rely heavily on evidence that is too high as chances are good that this evidence is not a reliable indicator of the criterion $y$. We therefore set the OWA weights to account for this data context by putting zero weight on the highest two data points with the remaining weights set to $1/(k-2)$. We acknowledge that both LWA and OWA use prior information in this simulation, the challange here is whether aggregators that combine the LWA and OWA can take advantage of this. In other words, the aim of the experiment is not to compare the operators in absolute terms, but to illustrate their individual behavior in reacting to this environment.

\begin{figure*}[ht]
\centering
  \includegraphics[width=\textwidth]{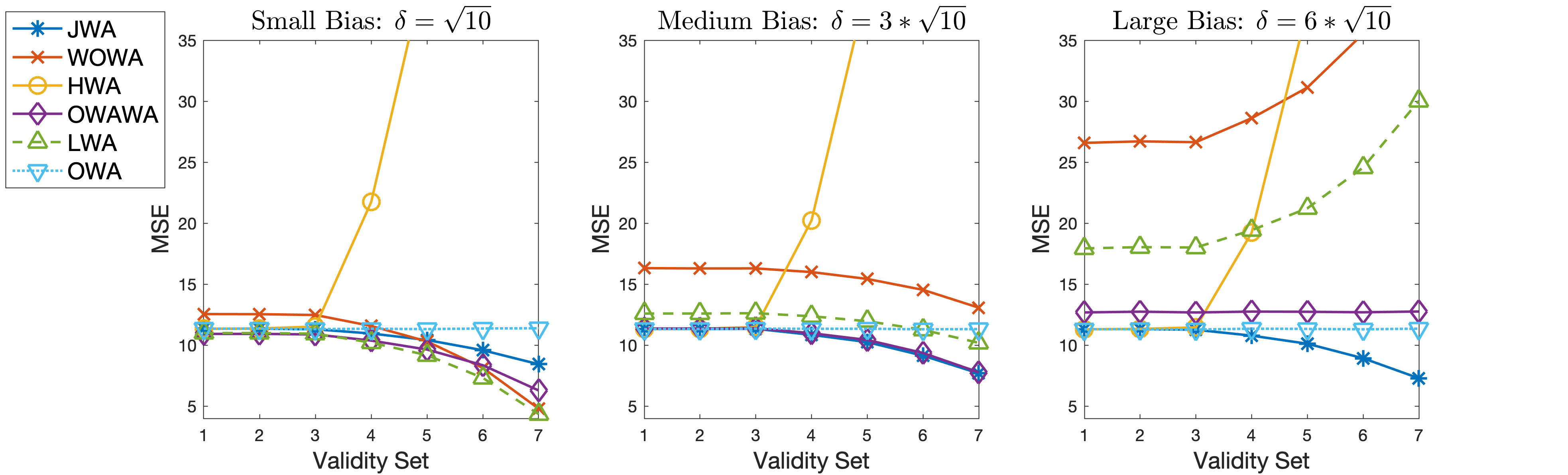}
\caption{Performance of operators in aggregating the evidence provided by 10 sources to predict a simulated criterion across different levels of bias and a series of data sets (x-axis), see Table \ref{tab:variable}. Note: continuous line for visibility only.}
\label{fig:EXP1}
\end{figure*}

Each aggregation operator is applied to the source data $X$ to generate an aggregate that represents a prediction $\hat{y}$ of $y$. We measure performance as the mean squared error (MSE) between $\hat{y}$ and $y$ across the $n$ trials. We vary the validity and bias parameters as shown in Figure \ref{fig:EXP1} and Table \ref{tab:variable}. We report the mean MSE across 500 replications for each parameter combination listed in Table \ref{tab:variable}.

In our results we show the performance of 4 different aggregation operators that combine the LWA and OWA: (a) JWA, (b) WOWA, (c) HWA, and (d) OWAWA (with $\alpha = 0.5$). For reference, we also show the the performance of the LWA and OWA. We simulate $X$ using the different sets of validity values for the sources displayed in Table \ref{tab:variable}. For each set of validity values, the LWA weights are equal to these validity values divided by their sum (so they sum to one). Therefore, the LWA operator weights sources by their validity, equally weighting them when the validity of each source is the same, and thus increasing weight on the higher validity sources as the data sets go from 1 (top row of Table \ref{tab:variable}) to 7 (bottom row in the top half of Table \ref{tab:variable}).

Figure \ref{fig:EXP1} displays plots of the MSE of each aggregator by the validity sets in Table \ref{tab:variable}. Each panel reflects a different level of bias as defined by the panel titles. From left to right panel, the bias increases from a small value (within 1 standard deviation of the data in $X$) to larger values that can more easily stand out relative to the unbiased sources. We can see that the LWA tends to increase performance as the difference between source validity values increase, but only when the bias term $\delta$ is not large. When $\delta$ is large, LWA performs worse as the validity sets increase. This is because the LWA effectively focuses on fewer \emph{valid} sources when the sources differ more, and these few valid sources can easily become completely unreliable when they are biased by a large $\delta$. The OWA is completely unaffected by the changes in both the validity sets and the bias size. This reflects how this aggregator is designed, equally weighting all pieces of evidence, except disregarding the largest two. 

The behavior of the JWA aggregate capitalizes on merging the LWA and OWA strategies, showing improved performance as the validity sets increase, yet remaining robust to the size of the bias that alters the data context. Indeed, JWA is able to leverage the LWA and OWA weights to produce a robustly good performance across all bias levels and validity sets. The OWAWA aggregator (which simply averages the output of LWA and OWA) outperforms JWA in the left panel, performs similarly to JWA in the middle panel, but performs far worse than JWA in the right hand panel. The left hand panel is a very friendly environment for the LWA, thus benefiting the OWAWA. However, the right hand panel is where the LWA's performance deteriorates due to the harsh data context. 

The WOWA and HWA, that allow the weights to interact, appear to perform worst than the OWAWA in all cases except validity sets 6 and 7 in the small bias panel. The HWA deteriorates quickly as the validity sets increase, while the WOWA deteriorates quickly as the magnitude of the bias increases.

Interestingly, the JWA and OWA do not change substantially across panels because we selected order weights that would be robust to this specific change in data context. While the LWA performance deteriorates, the JWA does not because it relies on the interaction between the LWA and OWA strategies to capitalize on information about source validity while still remaining robust to the random injection of bias. Indeed, the JWA shows subtly increased performance as bias increases, as this facilitates the identification and exclusion of the compromised sources, while benefiting from accurate weights for the sources remaining.

\begin{table}
\addtolength{\tabcolsep}{-1.5pt}
\centering
\caption{Variable Simulation Parameters}
\begin{tabular}{cccccccccccc} \hline
 & \multicolumn{10}{c}{Validity $\sigma_{x_i,y}$} \\
Set  & $x_1$ & $x_2$ &$x_3$ &$x_4$ &$x_5$ &$x_6$ &$x_7$ &$x_8$ &$x_9$ &$x_{10}$\\
  \hline
1 & \mbox{ }1.00\mbox{ } 	&	\mbox{ }1.00\mbox{ } 	&	\mbox{ }1.00\mbox{ } 	&	\mbox{ }1.00\mbox{ } 	&	\mbox{ }1.00\mbox{ } 	&	\mbox{ }1.00\mbox{ } 	&	\mbox{ }1.00\mbox{ } 	&	\mbox{ }1.00\mbox{ } 	&	\mbox{ }1.00\mbox{ } 	&	\mbox{ }1.00	\\
2 & 0.91	&	0.95	&	0.97	&	0.99	&	1.01	&	1.02	&	1.03	&	1.04	&	1.04	&	1.05	\\
3 & 0.68	&	0.80	&	0.89	&	0.96	&	1.01	&	1.06	&	1.10	&	1.14	&	1.17	&	1.20	\\
4 & 0.03	&	0.10	&	0.23	&	0.42	&	0.65	&	0.94	&	1.27	&	1.66	&	2.10	&	2.60	\\
5 & 0.00	&	0.01	&	0.03	&	0.10	&	0.25	&	0.51	&	0.95	&	1.62	&	2.59	&	3.95	\\
6 & 0.00	&	0.00	&	0.00	&	0.00	&	0.02	&	0.10	&	0.34	&	1.00	&	2.57	&	5.96	\\
7 & 0.00	&	0.00	&	0.00	&	0.00	&	0.00	&	0.00	&	0.03	&	0.23	&	1.52	&	8.22	\\
\hline
\end{tabular}
\label{tab:variable}
\end{table}

\section{Discussion}

We have introduced a conceptual framework that recognizes the LWA and the OWA as operationalizations of two independent strategies for aggregating data. We consider these strategies to be represented by their respective weights, and build a mathematical framework that appropriately treats the weights of each strategy as parts of a whole in order to blend the strategies. This leads to the use of compositional geometry to blend weights that represent the quality of sources and the quality of the evidence they produce. We have shown that prior work on information aggregation does not jointly account for both the quality of source and the quality of evidence. Through the examples in Table \ref{tab:example} and Figure \ref{fig:EXP1}, we highlight the importance of simultaneously considering both the quality of sources and the quality of information arising from said sources, showing how these can be independently defined and leveraged to improve information aggregation and the handling of inter-source uncertainty. We introduce the Joint Weighted Average (JWA) as the first aggregator capable of jointly and systematically focusing on these two features that define the quality of evidence. 

The JWA jointly focuses on source-and-evidence specific worth by directly combining the linear and ordered weights using compositional geometry \cite{Aitchison1982,smithson2022compositional,PekaslanCompositional2021}. This approach allows researchers more transparency and a strong mathematical framework to understand the JWA's properties and to augment the JWA for specific applications. Viewing the weights from LWA and OWA as compositions also naturally leads to better control and understanding of the combined weights allowing more flexibility than previously proposed approaches.

This approach reveals how previous work attempting to combine LWA and OWA have suggested that a combined opperator must adhere to a set of mathematical properties that precludes this joint focus. Specifically, \cite{SDOWA_Cardin2021,llamazares2011generalizations} propose that the output of aggregators that combine LWA and OWA must be bound between the outputs of LWA and OWA. We show that this property does not allow the strategies of LWA and OWA to be fully merged, and the cost of this property is clearly shown in the right hand panel of Figure \ref{fig:EXP1}. Averaging the outputs of LWA and OWA leads to poor performance if one of the two aggregates faces an environment that degrades their approach, where the JWA can withstand such environments by blending their strengths more effectively.

\subsection{Conclusions and Future Work}

The JWA represents a single, simple, interpretable instance of a potentially broad class of weighting strategies that combine individual weighting strategies based on the quality of sources and the quality of evidence. By articulating the compositional nature of this type of combination, and leveraging the well-defined framework of compositional geometry, the JWA operator enables the combination of aggregation strategies in a systematic fashion which has thus far been out of reach. 

This provides on the one hand new pathways to understanding how humans combine information and make decisions, and on the other, a mechanism to aggregate information in applications from healthcare to mine-detection, and machine-learning more broadly. Perhaps the most exciting potential lies with the JWA providing a bridge linking the processes underpinning human and artificial reasoning, in turn offering more human-centric, understandable, and explainable approaches to AI.

In view of this, we plan to explore how this aggregator might be fit to data to estimate the relative contribution of linear and ordered weights in unknown systems that generate aggregates, such as human judgment--in particular under uncertainty. We believe that this aggregator has the potential to describe psychological processes in a novel way, as well as providing a tool that could automate future aggregations made in this way. 

Second, the linear and order weights can be elicited from human users or experts numerically, or indeed as intervals or fuzzy sets--capturing associated uncertainty, to generate an automated information aggregation system with well known mathematical properties. In the above examples, we provided weights for the sources and the evidence that reflected priorities for aggregates. As such the JWA can be used to define what information would need to be elicited as well as the implications for what aggregates these two sets of weights would produce, to see if human users agree that the resulting aggregates indeed reflect the priorities given.

Finally, combining linear and order weights plays an important role in XAI by providing parameters that allow people to understand the role of the source vs. the role of evidence in aggregation and act upon it. More broadly, at the level of AI and machine learning, where LWA approaches are used pervasively, all the way to the heart of neural networks; broadening the perspective and research to explore joint source and evidence weights holds substantial promise.

\subsubsection*{Acknowledgements}
The research underpinning this work was in part supported by the NSF (SES-2343580) and UKRI (ES/Z000084/1) co-funded project \emph{A Novel Theory of Ordered Judgment Processes}.

\bibliography{ref, smps-template}
\bibliographystyle{splncs04}

\end{document}